\documentclass[runningheads]{llncs}
\usepackage[T1]{fontenc}
\usepackage{graphicx}
\usepackage{enumitem}
\usepackage{amsmath}
\usepackage[colorinlistoftodos]{todonotes}
\usepackage{comment}

\usepackage[ruled,linesnumbered]{algorithm2e}
\DontPrintSemicolon 

\usepackage{tabularx}     
\usepackage{booktabs}     
\usepackage{caption}      
\usepackage[caption=false]{subfig} 

\begin{document}

\title{Toward Simple and Robust Contrastive Explanations for Image Classification by Leveraging Instance Similarity and Concept Relevance}

\titlerunning{Toward Simple and Robust Contrastive Explanations}
\author{Yuliia Kaidashova\inst{1}\orcidID{0009-0005-9084-4796} \and
Bettina Finzel\inst{1}\orcidID{0000-0002-9415-6254} \and
Ute Schmid\inst{1}\orcidID{0000-0002-1301-0326}}
\authorrunning{Y. Kaidashova et al.}
\institute{Cognitive Systems, University of Bamberg, Weberei 5, 96047 Bamberg, Germany
\email{\{ute.schmid,bettina.finzel\}@uni-bamberg.de}\\
\email{yuliia.kaidashova@stud.uni-bamberg.de}\\
\url{https://www.uni-bamberg.de/kogsys/}
}

%
%
%
\maketitle              
\begin{abstract}
Understanding why a classification model prefers one class over another for an input instance is the challenge of contrastive explanation. This work implements concept-based contrastive explanations for image classification by leveraging the similarity of instance embeddings and relevance of human-understandable concepts used by a fine-tuned deep learning model. Our approach extracts concepts with their relevance score, computes contrasts for similar instances, and evaluates the resulting contrastive explanations based on explanation complexity. Robustness is tested for different image augmentations.
Two research questions are addressed: (1) whether explanation complexity varies across different relevance ranges, and (2) whether explanation
complexity remains consistent under image augmentations such as rotation and noise. The results confirm that for our experiments higher concept relevance leads to shorter, less complex explanations, while lower relevance results in longer, more diffuse explanations. Additionally, explanations show varying degrees of robustness. The discussion of these findings offers insights into the potential of building more interpretable and robust AI systems.

\end{abstract}
\section{Introduction}

Artificial Intelligence systems have produced impressive results because of the rapid development of deep learning, which has resulted in significant technological advancement and broad adoption. But a big problem with deep learning models is that they frequently act as ``black boxes''. This may result in an inability to evaluate such systems' reliability, endangering adoption by human users due to a lack of trust in AI \cite{adadi2018peeking,ali2023explainable,bruckert2020roadmap,gunning2019darpa,rudin2019stop}.

The goal of Explainable Artificial Intelligence (xAI) is to shed light on the inner workings of these ``black-box'' models \cite{gunning2019darpa,vanLent2004explainable}. 
Among the diverse range of xAI approaches \cite{adadi2018peeking,ali2023explainable,schwalbe2023comprehensive,stepin2021survey}, \textit{concept-based explanations} (see, e.g., Poeta et al. \cite{poeta2023concept}) currently stand out for their focus on interpretability. We introduce \textit{concept-based contrastive explanations} that explain a model's decision using human-understandable concepts that distinguish instances from different classes.

As the human is considered in this work as the recipient of explanations, we refer to the term \textit{concept} from a cognitive psychology perspective. A concept is a mental representation that helps people categorize and name objects, events, or ideas based on shared features \cite{margolis2007ontology,palmer2002psychological}. Concepts further play a crucial role in cognition by allowing humans to recognize patterns, make decisions, and infer the properties of new instances based on prior knowledge \cite{margolis2007ontology,palmer2002psychological}.

One important aspect related to a human's processing of information is the evaluation of an explanation's complexity (see, e.g., Schwalbe \& Finzel for an overview of methods and metrics \cite{schwalbe2023comprehensive}). As demonstrated by Kulesza et al. \cite{kulesza2023too} an explanation should be as concise and complete as possible, given that it contains the relevant information to be well understood and processed by human users.

In this work, explanation complexity is measured by the length of an explanation which is determined by the number of concepts included. The shorter (simpler) explanations are preferred, as they likely improve human understanding in cases that may be easily confused otherwise \cite{finzel2024near,winston1975learning}.

Another important criterion is explanation robustness, which refers to the stability of an explanation when the input is slightly modified \cite{ali2023explainable,9176802,schwalbe2023comprehensive}.
According to Arcaini et al., a common way to test robustness is image augmentations, which are small, controlled modifications applied to input images \cite{9176802}. Gaussian noise is one of the typical augmentation techniques, where adaptations are normally distributed across the image to simulate sensor noise. Image rotation is very common, too (e.g., by 1 to 180 degrees to test the model’s consistency under geometric transformations) among other techniques such as brightness variation, blur, and compression \cite{9176802}.
Robust explanations should remain consistent across similar inputs, maintaining the same set of concepts and preserving the explanation length \cite{9176802}. This work focuses on alterations by introducing Gaussian noise and by applying either 10 degrees or 180 degrees rotation to input images as all these modifications are expected to impact the recognition of material and concept-specific spatial orientation \cite{9176802}.

Further, the domain of image classification is a suitable use case for investigating and improving the interpretability of learned models with concept-based contrastive explanations. Even though models like Convolutional Neural Networks (CNNs) are found to be very accurate, they often make decisions based on small details and diverse pixel values, which makes it difficult to understand why they chose a certain label for an image. Concept-based contrastive explanations can help by showing which visual parts of the image were important for the model’s decision, and how human-understandable representations of such visual parts made the model choose one class over another. Linking these concepts and contrasts with the relevance of each visual part, allows for an explanation that takes into account the different contribution strengths of concepts that result from the model's weight learning.

To explore the potential of concept-based contrastive explanations in image classification, this work focuses on explanation complexity and robustness and considers different relevance ranges for explanation generation. More specifically, relevance ranges are determined based on quantiles over the relevance distribution of relevance-ranked concepts. Accordingly, the following research questions are formulated for the evaluation of our novel approach:

\begin{enumerate}[label=\textbf{R\arabic*}, topsep=0pt]
    \item \label{rq:r1} Does the length of concept-based explanations vary for similar images within different relevance ranges?
    \item \label{rq:r2} Does the length of concept-based explanations remain stable under image augmentations such as noise, 10-degree rotation, and 180-degree rotation?
\end{enumerate}

\noindent
Subsequently, we introduce the related work that inspired our approach (Section \ref{sec:related}), introduce our methodology (Section \ref{sec:method}), evaluate and discuss it (Section \ref{sec:evaluation} and Section \ref{sec:discussion}) and conclude with a summary of our findings in Section \ref{sec:conclusion}.

\section{Related Work}\label{sec:related}

Linking cognitive research with xAI, the term \textit{concept} typically refers to higher-level, semantically meaningful attributes learned by a model, which can be interpreted and understood by humans \cite{poeta2023concept}. In the explainability literature, \textit{concepts} are defined in various ways, including, but not limited to symbolic concepts, unsupervised concept bases, prototypes, and textual concepts \cite{finzel2024relation,poeta2023concept}. This work focuses on symbolic concepts, which are human-defined symbols representing interpretable abstractions of a feature, such as \textit{red}, \textit{striped} or \textit{handle} \cite{finzel2024relation,poeta2023concept}. Such concepts are well-suited for model interpretability since they are naturally consistent with human understanding.

To fully leverage the usefulness of concepts in explaining model predictions, it is essential to understand what concept-based explanations are. One of the categories of concept-based explanations is the \textit{node-concept association}, which refers to explicitly linking internal units (e.g., neurons or filters) in a neural network to specific concepts \cite{finzel2024relation,poeta2023concept}. This can be achieved either post hoc, by analyzing which concepts maximally activate certain units, or during training, by requiring a unit to predict a concept \cite{poeta2023concept}. This approach improves model transparency by revealing whether and which internal representations correspond to human-understandable terms, thereby allowing explainees to better understand what a network has learned \cite{achtibat2023attribution,finzel2024relation,poeta2023concept}.

Contrastive explanations aim to explain why a model chose one class over another by contrasting outcomes for instances a model has been applied to (see, e.g., Stepin et al. \cite{stepin2021survey}). Unlike traditional explanations that focus on a single class, contrastive explanations identify which features would need to be altered in order to predict a different class or which features are missing to support the alternative class. In the context of xAI, understanding how humans naturally interpret explanations is crucial \cite{bruckert2020roadmap,doshivelez2017rigorous,rudin2019stop}. Research in philosophy and cognitive science highlights that people do not typically ask for explanations in isolation ``Why P?'', but rather in relation to an alternative that did not occur (see, e.g., Miller \cite{miller2019explanation}). This form of contrastive explanation is typically structured as ``Why P rather than Q?'', where \textit{P} is the observed fact and \textit{Q} is the imagined foil, an alternative that was expected or considered but did not occur. The foil may not be explicitly stated in the question, but humans are generally capable of inferring it from context, such as tone, expectations, or background knowledge \cite{miller2019explanation}. One of the very first contrastive approaches was introduced, e.g., in Winston's seminal work on \textit{near misses} that are differently classified explanatory examples lacking a minimal set of properties. \cite{rabold2022nearmisses,winston1975learning}.

Lipton emphasizes that a contrastive explanation is often cognitively easier to produce and more relevant to human reasoning than exhaustive explanations of a fact \cite{lipton1990contrastive}. For example, when classifying an image as a teapot rather than a vase, mentioning a ``spout'' might be sufficient to explain the contrast, even if it does not completely explain why some vessel is a teapot (see Finzel et al. \cite{finzel2024relation}).

Building on the theoretical foundations of contrastive explanation, several approaches in machine learning have proposed methods to generate such explanations for neural networks. One known area of research is using input space perturbations to create contrastive explanations \cite{poeta2023concept,stepin2021survey}. For example, the work of Dhurandhar et al. \cite{dhurandhar2018explanations} introduced the concept of \textit{pertinent negatives}, which are features that are deemed irrelevant by a model and need to be included to get another prediction. Their method generates contrastive explanations by identifying minimal changes to an input that would alter the classification, effectively answering the question: ``Why is input x classified in class y?''. This is achieved by finding points close to the original input that lie just outside the decision boundary of the current class but within the boundary of a contrasting class. These points help highlight what is missing in the input to prevent classification into the foil class, thus providing a contrastive explanation through absence.

Prabhushankar et al. \cite{prabhushankar2020contrastive} propose a method for contrastive explanations in neural networks that focuses on identifying salient features that separate two classes. Their approach constructs explanations by comparing the internal activations of the neural network for two target classes: the predicted class and an alternative class. By analyzing differences in feature importance between these classes, the method highlights the contrastive attribution, that is the set of features that pushes the model towards one class and away from another.

While both methods (\cite{dhurandhar2018explanations} and \cite{prabhushankar2020contrastive}) effectively identify differences between classes, they primarily focus on plain features rather than human-understandable concepts. This can limit the interpretability for users, as the explanations lack semantic clarity. Including recognizable concept-based elements could significantly improve human perception and trust in the explanations.

Recent works by Finzel et al. \cite{finzel2024relation,finzel2024near} propose a framework for contrastive explanations based on the extraction of human-understandable concepts from node-concept associations to explain model decisions in image classification. The concept-based and relational explainer introduced in \cite{finzel2024relation} utilizes a sampling method that selects concepts for explanation generation based on quantiles over the relevance distribution of concepts, where concepts are ranked by the strength of their relevance. The conceptual framework presented in \cite{finzel2024near} suggests to compute the similarity of \textit{embeddings} (a model's internal vectorized representation of input data) for the generation of concept-based contrastive explanations as embeddings can yield more meaningful similarity scores compared to plain input. 

This work presents an actual implementation of these ideas utilizing embeddings for similarity computation and adds an evaluation of the robustness and interpretability of \textit{contrastive} explanations dependent of different \textit{relevance ranges} (different quantiles over the relevance distribution) for node-concept associations. It builds upon the state of the art in concept-based and contrastive explanation and particularly contributes an evaluation based on measurable metrics for robustness and an empirical investigation of the relevance range's effect on explanation simplicity.

\section{Concept-based Contrastive Explanations}\label{sec:method}

Our methodology for generating human-understandable contrastive explanations is applied to image classification with a CNN. The model is based on the \textit{VGG16} architecture as introduced by Simonyan \& Zisserman \cite{simonyan2014very}, in particular a CNN pre-trained on ImageNet. CNNs consist of convolutional layers that automatically extract visual features from images, followed by classification layers that use these features to predict class labels.

For this research, a publicly available subset of \textit{ImageNet} was used (see Deng et al. \cite{deng2009imagenet}) called the \textit{ImageNet Large Scale Visual Recognition Challenge} (ILSVRC) 2012-2017 image classification and localization dataset (published by Russakovsky et al.  \cite{ILSVRC15}).
ILSVRC uses \textit{WordNet} from Princeton University \cite{miller1995wordnet} to construct a semantic hierarchy that helps in selecting similar classes that still have meaningful distinctions. For this work, we selected indirect child classes of a common parent node to enable reasonable contrastive explanations, ensuring that the classes share some visual similarities while still being distinct. In particular, we set up the data and model to perform classification between \textit{teapot} and \textit{vase} images. Contrasting these classes is a cognitively interesting case, due to their similarity, yet, heterogeneity in design \cite{finzel2024relation}. Following the \textit{WordNet} hierarchy, the path is as follows: \textit{artifact $\xrightarrow{}$ instrumentality $\xrightarrow{}$ container $\xrightarrow{}$ vessel $\xrightarrow{}$ teapot/vase}. The number of images per class varies slightly. For fine-tuning the CNN model (with 2 output neurons), we used 622 images of teapots and 636 images of vases. The test split includes 678 teapot images and 663 vase images.

\begin{figure}[htb!]
	\centering
	\includegraphics[width=0.95\textwidth, alt={Flow chart illustrating a machine learning process for concept-based contrastive explanation. The process begins with a train set (D) and a test set (D'). A fine-tuned model (M') extracts embeddings from the test set. Target and set of contrastive embeddings are selected. Cosine similarity (CosSim) is calculated to find the best match. Concept extraction and ranking lead to the formation of concept sets (C1, C0). A histogram shows ranges of values (0\%-25\%, 25\%-50\%, 50\%-75\%, 75\%-100\%), and after naming process contrastive explanation (E) generated with specified length. Icons represent concepts like teapots and flowers.}]{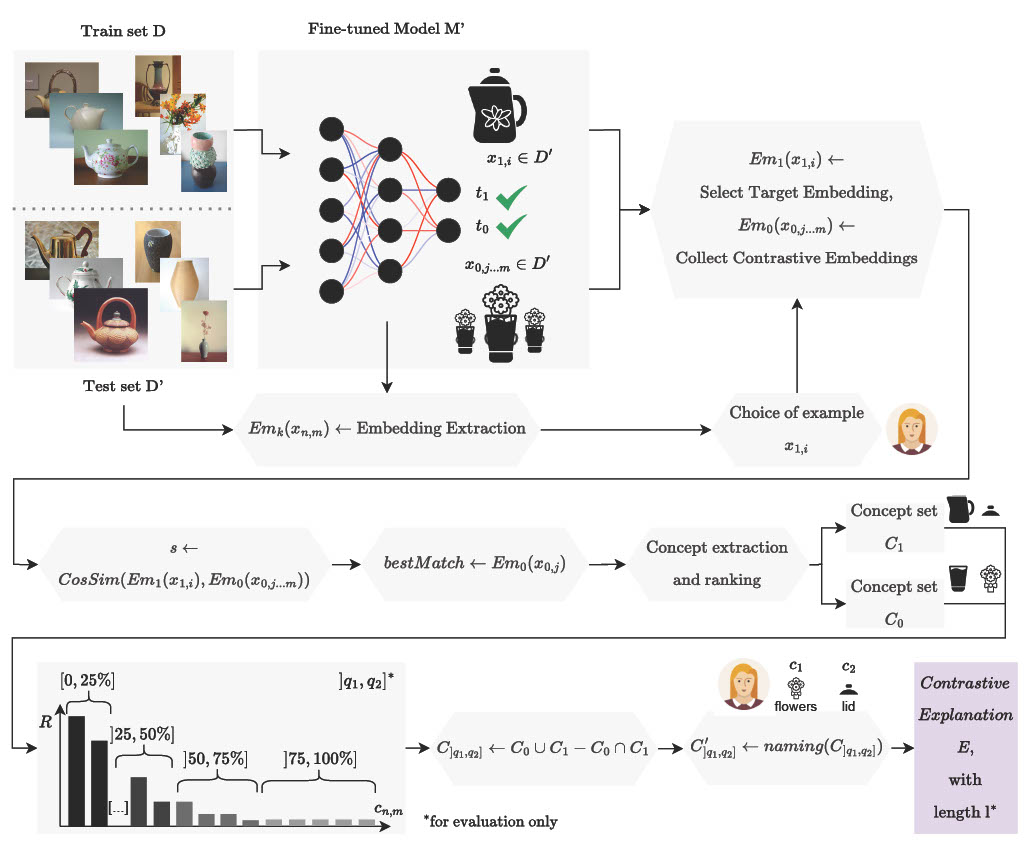}
	\caption{Illustration of the overall concept-based contrastive explanation approach following the steps of Algorithm \ref{alg:1} and extended by the mention of relevance ranges and explanation length as evaluation variables (best viewed in color).}
	\label{fig:overview}
\end{figure}

To generate contrastive explanations that highlight the distinguishing features between visually and semantically related classes, Algorithm \ref{alg:1} was applied (see also its illustration in Figure \ref{fig:overview}).

First, the model $M$ is fine-tuned on instances from the training data $D$ belonging either to a target class $t_1$ or the contrastive class $t_0$ (e.g., teapot vs. vase) to prepare the explanation generation for the fine-tuned model $M'$ (see Algorithm \ref{alg:1}, line 8). Second, embeddings ($Em_0$ or $Em_1)$ are extracted for instances from the test set $D'$ that are correctly classified by $M'$, utilizing only feature extraction layers of $M'$ (line 9). Third, an example's embedding $Em_1(x_i)$ is selected, either by choosing the image instance $x_i$ arbitrarily or by human choice (line 10).

\begin{algorithm}[H]
      \caption{Concept-based Contrastive Explanation} \label{alg:1}
    \scriptsize \textbf{Input:}\;
    \scriptsize $(M, D, D', C_i)$, where $M$ is the model, $D$ is the training set, $D'$ is the test set, and $C_i$ is the set of concepts for each class ($t_1$ for the target, $t_0$ otherwise) \;
    \scriptsize \textbf{Output:}\;
    \scriptsize  $E$: concept-based contrastive explanation \;
    \scriptsize \textbf{Begin:}\;
    
    \Indp 
    \scriptsize $\text{maxSim} \leftarrow -1$\;
    $\text{bestMatch} \leftarrow \emptyset$\;
    $M' \leftarrow FineTune(M, D)$\;
    $Em_{\text{0}}, Em_{\text{1}}  \leftarrow embedding(M', D')$, embeddings of correctly classified instances\;
    $Em_1(x_i) \leftarrow selectExampleEmbedding(Em_1)$\;
    \scriptsize \textbf{For each} $Em_0(x_j) \in Em_{\text{0}}$ \textbf{do}\;
    \Indp
      \scriptsize $s \leftarrow CosSim(Em_0(x_j), Em_1(x_i))$, where $x_i$ of class $t_1$, $x_j$ instance of class $t_0$ \;
      \scriptsize \textbf{If} $s > \text{maxSim}$ \textbf{then}\;
      \Indp
        \scriptsize $\text{maxSim} \leftarrow s$\;
        $\text{bestMatch} \leftarrow x_j$\;
      \Indm
      \Indm
        
      \scriptsize $C_1 \leftarrow ranking(extract(x_i, M', D'))$\;
      \scriptsize $C_0 \leftarrow ranking(extract(bestMatch, M', D'))$\;

      \scriptsize $E \leftarrow contrastive(naming(difference(C_1, C_0)))$
      
\scriptsize  \textbf{Return} $E$\;
\scriptsize \Indm \textbf{End}
\normalsize
\end{algorithm}

\noindent

\vspace{0.2cm}

\noindent
Fourth, the cosine similarity is computed between the embedding of the input image $Em_1(x_i)$ and the embeddings of all images from the opposite class $Em_0(x_j)$. The best matching contrastive image $x_j$ is the one with the highest cosine similarity between embeddings, that is the most semantically and visually similar image to $x_i$ from the opposite class (lines 11-15).
Next (see lines 16-17), concept extraction is performed by capturing internal relevance attributions from predefined convolutional layers (e.g., layer 40) with the help of Concept Relevance Propagation (CRP).
CRP as introduced by Achtibat et al. \cite{achtibat2023attribution} links neuron-level signals derived during the forward pass to concepts by performing a masked backward pass. To isolate a specific concept, CRP selectively masks irrelevant pathways during propagation, allowing only the neurons associated with that concept to retain relevance. This yields concept-specific relevance scores that quantify how much each concept (neuron) contributed to the prediction.
In our approach, all concepts are sorted in descending order based on their contribution to the prediction (also lines 16-17).
In particular, two concept sets ($C_1$ and $C_0$) are constructed, one for instances of $t_1$ and one for $t_0$. Next, unique concepts are identified, defined as those that occur in only one sample from either class. Concepts that appear in both classes are excluded to ensure that the contrastive explanation is focused on class-distinctive features. This step is coupled with optional naming (automated or manual) of features to turn them into human-understandable concepts (line 18).
The final explanation is formatted like so:

\begin{quote}\label{form:contr-expl}
\small
The model classified the image as a \textbf{Target class} instead of a \textbf{Contrastive class} because it contains the concepts \textit{c\textsubscript{1,1}, c\textsubscript{1,2}, ..., c\textsubscript{i,n}}, and does not contain the concepts \textit{c\textsubscript{0,1}, c\textsubscript{0,2}, ..., c\textsubscript{j,m}}.
\normalsize
\end{quote}

\begin{figure}[htb!]
	\centering
	\includegraphics[width=0.92\textwidth, alt={Figure illustrating exemplary concepts and resulting concept-based contrastive explanation. The top row shows images of a teapot labeled 1 to 4, with concepts like handle and spout. The bottom row displays images of a vase labeled 5 to 8, with concepts like tapering bottom and neck. A text box explains the model's classification of the image as a teapot due to the presence of certain concepts and absence of others. Keywords: teapot, vase, classification, model, handle, spout, tapering bottom, neck.}]{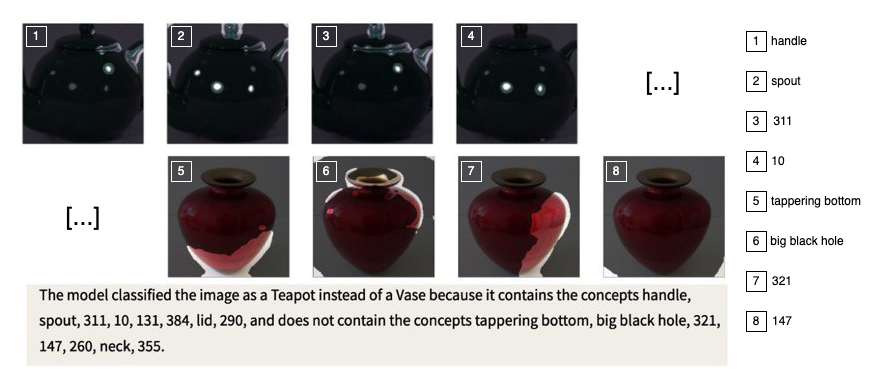}
	\caption{Contrastive explanation for teapot and vase (best viewed in color)}
	\label{fig:contr-expln}
\end{figure}

\noindent
Here, $c\textsubscript{i,n}$ refers to the unique concepts found in the target class, while $c\textsubscript{j,m}$ corresponds to the unique concepts found in the contrastive class. For a concrete example, see Figure \ref{fig:contr-expln}. The contrast helps to understand why one class is preferred over another and to see the main concept differences between classes\footnote{Note that optional concept naming is supported by techniques such as reference sampling based on activation maximization or similar methods \cite{achtibat2023attribution}. Unnamed concepts may shed light on ambiguities or complexities in finding a suitable name. Otherwise, the model needs to be provided with pre-defined concepts \cite{poeta2023concept}.}.

\noindent
For the evaluation, the sorted concepts are grouped into categories representing different levels of relevance, from most to least influential. These relevance levels form the basis for evaluating explanation length and robustness. Explanation lengths are compared across the different relevance levels to examine whether highly relevant concepts tend to produce shorter and more focused explanations than less relevant ones (see next Section for a detailed description).

\section{Evaluation and Results}\label{sec:evaluation}
The goal of \ref{rq:r1} is to examine whether the length of concept-based explanations differs depending on the relevance strength of the concepts. To address this, explanations are analyzed in distinct relevance ranges.
To compute the ranges, all concepts are first sorted in descending order of relevance for a given prediction, where \textit{very strong} relevance is the top 25\% of total relevance, \textit{strong} are concepts that lie between 25\% and 50\%, \textit{low} are concepts that account for the 50\% to 75\% and \textit{very low} is the remaining 25\% of the cumulative relevance.
The evaluation of robustness for \ref{rq:r2} is based on comparing the original images and their augmented versions within relevance ranges, assessing whether the length of concept-based explanations remains stable.
We formulate the following hypotheses, accordingly:

\vspace{0.2cm}

\begin{enumerate}[label=\textbf{H\arabic*}, topsep=0pt]
    \item \label{h:h1} Regarding explanation length, it is expected that similar images with concepts of \textit{very strong} relevance lead to shorter explanations compared to concepts with \textit{strong}, \textit{low}, and \textit{very low} relevance. Furthermore, it is expected that concepts with \textit{strong} relevance result in longer explanations than those with \textit{very strong} relevance, but shorter explanations than concepts with \textit{low} and \textit{very low} relevance. Similarly, concepts with \textit{low} relevance are expected to produce longer explanations than those with \textit{very strong} and \textit{strong} relevance, but shorter explanations than those with \textit{very low} relevance. Finally, concepts with very \textit{low} relevance are expected to lead to the longest explanations compared to all other relevance levels.
    \item \label{h:h2} Regarding robustness, it is expected that within the relevance ranges the explanation length remains consistent between the original explanation (produced using the unaltered image) and the explanation generated after applying an augmentation technique to the input image (e.g., noise, rotation).
\end{enumerate}

\begin{figure}[htb!]
    \centering
    \includegraphics[width=0.6\linewidth, alt={Box plot showing explanation length across four categories: very strong, strong, low, and very low. Each category displays a box with median line, interquartile range, and whiskers indicating variability. Outliers are marked as individual points. Explanation length increases from very strong to very low.}]{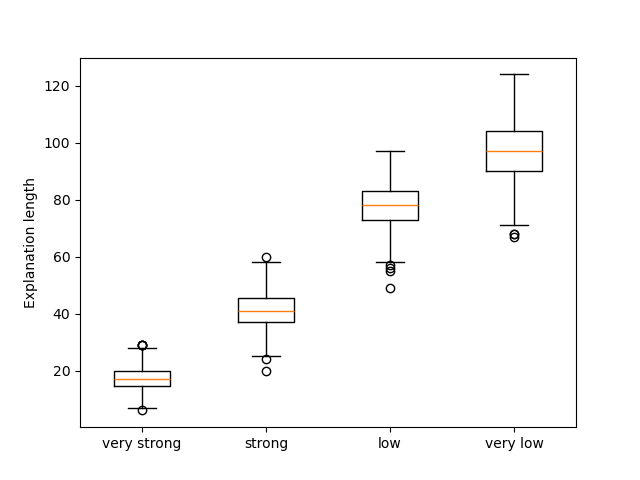}
    \caption{Explanation length across relevance ranges (n = 615 teapots, 328 vases)}
    \label{fig:expl-length}
\end{figure}

\noindent
The evaluation of explanation length for \ref{rq:r1} is performed with an ANOVA test as introduced by Fisher \cite{fisher1992statistical} that assesses whether the means of multiple groups differ significantly, where the four relevance ranges are the test groups. 
We can see in Figure \ref{fig:expl-length} that the explanation length varies depending on the relevance range.
The maximum possible explanation length is 512, which would occur if one class contained all 512 concepts while the other contained none\footnote{The number of extractable concepts depends on the architecture of the model in use.}.

\noindent
For \textit{very strong} relevance, the median explanation length is around 17. The length in the 25th to 75th percentile ranges from 14 to 21. For some explanations the length ranges from 8 to 30. Only a few explanations have a length below 10 and above 30.
Considering the \textit{strong} relevance, the median explanation length is around 40. Most values lie between 35 and 45, while overall, the explanation length from this group mainly varies from 25 to 60, with some outlier samples having a length below 25 and above 60.
For \textit{low} relevance, the median explanation length increases to 78, with most explanations ranging from 70 to 85. The length mainly varies from 55 to 98, with only few outliers having a length below 55. 
Considering the \textit{very low} relevance, we can see that the median value of explanation length is around 100. For most samples the explanation length is between 90 and 110, with a general variation between 70 to 123 and some outliers below 70.
Also, the results of the ANOVA test show a high F-value of 354.41 and a statistically significant p-value below 0.001.

The robustness evaluation for \ref{rq:r2} is performed using a paired t-test as introduced by Student \cite{student1908probable}, comparing the means of sampled explanation lengths for original images and their augmented versions (Rotation 180°, Rotation 10°, Gaussian Noise) for each relevance range (\textit{very strong}, \textit{strong}, \textit{low}, and \textit{very low}).

For the augmentation of 180-degree rotation, we can see in Figure \ref{fig:rot} (Appendix) that the distribution of explanation lengths is visually quite similar between original and rotated images across all relevance levels. The paired t-tests (see Table \ref{tab:t-test-augm}) support this observation: For \textit{very strong}, \textit{strong}, and \textit{low} relevance, the differences are not statistically significant (p > 0.05). Only the \textit{very low} relevance category shows a significant difference (p = 0.0002), where rotated images tend to have slightly longer explanations.

For the augmentation of 10-degree rotation, Figure \ref{fig:rot10} (Appendix) reveals more noticeable differences. For \textit{very strong} relevance, explanation length appears slightly reduced for rotated images, though not significantly. However, for \textit{strong} and \textit{low} relevance, explanation lengths are significantly shorter after rotation (p = 0.0006 and p = 0.0175 respectively). For \textit{very low} relevance, no significant difference is observed (p = 0.7401), with nearly identical distributions.

For the augmentation with Gaussian noise, the effect of adding it is especially evident in Figure \ref{fig:gausnois} (Appendix). Across all relevance levels, explanation lengths are consistently and significantly lower for noisy images compared to the original ones. This is confirmed by the very low p-values (all < 0.001). The boxplots clearly show downward shifts in the distributions, most noticeably in the \textit{very strong} and \textit{strong} relevance categories.

\begin{table}[htb!]
\scriptsize
    \centering
    \caption{Results of t-tests for different augmentations across relevance ranges}
    \begin{tabularx}{\textwidth}{X | cc | cc | cc}
        \toprule
        \textbf{Relevance Range} 
        & \multicolumn{2}{c|}{\textbf{180° Rotation}} 
        & \multicolumn{2}{c|}{\textbf{Gaussian Noise}} 
        & \multicolumn{2}{c}{\textbf{10° Rotation}} \\
        \cmidrule(r){2-3} \cmidrule(r){4-5} \cmidrule(r){6-7}
        & \textbf{t} & \textbf{p} 
        & \textbf{t} & \textbf{p} 
        & \textbf{t} & \textbf{p} \\
        \midrule
        Very strong & 1.7630 & 0.0784 & 25.7762 &\textbf{ 0.0000} & 5.6708 & \textbf{0.0000} \\
        Strong      & -1.9153 & 0.0559 & 18.1813 & \textbf{0.0000} & -3.4474 & \textbf{0.0006} \\
        Low         & 1.7312 & 0.0839 & 13.2428 & \textbf{0.0000} & -2.3825 & \textbf{0.0175} \\
        Very low    & 3.7505 & \textbf{0.0002} & 13.9907 & \textbf{0.0000} & -0.3318 & 0.7401 \\
        \bottomrule
    \end{tabularx}
    \label{tab:t-test-augm}
\normalsize
\end{table}

\section{Discussion}\label{sec:discussion}

The results derived from t-testing the effect of different augmentations, in combination with the outcomes of the ANOVA test, demonstrate a clear relationship between relevance strength and explanation length. Explanations that were based on highly relevant concept groups (e.g., \textit{very strong}) tended to be shorter, while explanations involving less relevant concept groups (e.g., \textit{very low}) were significantly longer.
This trend suggests that when a model identifies highly relevant concepts, explaining its decision can be done more briefly and efficiently. In contrast, the longer explanations can be seen as an indicator for contrastive explanations being less effective, when the included concepts received less relevance by the learned model. This also supports the initial claim that highly relevant concepts are more discriminative, fewer of them are needed to form a confident prediction and explanation. On the other hand, longer explanations in the \textit{very low} relevance group may reflect model uncertainty, concept redundancy, or the lack of a dominant set of concepts. These findings support \ref{h:h1}, as we clearly see that lower relevance was associated with longer explanations.

\ref{h:h2} stated that the explanation length would remain the same for original images and their augmented versions within a certain relevance range. However, the results showed that this assumption only holds partially.

For rotation by 180 degrees, which is a strong geometric transformation, the effect on explanation length was relatively small. Only the \textit{very low} relevance range showed a significant change, while the more important relevance levels remained stable. This suggests that the most relevant concepts used by the model were largely preserved even when the image was flipped upside down. It seems that the model benefited from a degree of orientation invariance, possibly due to highly relevant concepts not changing in appearance if flipped (such as a vertical handle). The slightly varying results in the lower relevance ranges might reflect small changes in concept activations. Overall, these findings mostly support \ref{h:h2} for this type of augmentation.

In contrast, rotation by 10 degrees, although much more subtle, led to more noticeable changes in explanation length. Significant differences in the \textit{very strong}, \textit{strong}, and \textit{low} relevance categories indicate that even slight misalignments can alter the model’s internal concept processing. Some explanations may become shorter, others longer, possibly depending on which concepts gain or lose relevance after the rotation. Unexpectedly, the \textit{very low} relevance group remained stable, which may imply that such minor transformations primarily affected the core discriminative concepts. These results suggest that even small augmentations can influence the structure of the explanation, which is challenging \ref{h:h2} in this case. Comparing both rotation cases, the results nevertheless confirm that robustness under augmentation highly depends on the geometrical properties of learned concepts.

For Gaussian noise, hypothesis \ref{h:h2} clearly does not hold. Explanation lengths increased significantly across all relevance ranges when noise was added. Although the semantic content of the image remained the same, the model reacted strongly to this low-level perturbation. It appeared to lose focus on a small set of discriminative concepts and instead activated a broader range of concepts, leading to longer and more diffuse explanations. This indicates a lack of robustness in how the used model handles noisy inputs and suggests that its concept-based reasoning becomes less certain under such conditions.

In summary, hypothesis \ref{h:h2} was partially supported. While explanation lengths remained mostly stable under strong global changes like 180° rotation, even slight rotations or noise could lead to substantial shifts in explanation structure. This shows that the produced contrastive explanations are somewhat sensitive to image augmentations, especially those affecting low-level visual details.

While the results offer valuable insights into the behavior of concept-based contrastive explanations, several limitations should be acknowledged.
First, the evaluation of explanation length was based on the assumption that shorter explanations are preferable, as they are generally easier for humans to interpret and understand. While this assumption is supported by prior work on explanation comprehensibility \cite{miller2019explanation}, explanation length alone may not fully capture the quality or faithfulness of an explanation \cite{schwalbe2023comprehensive}. A short explanation might omit relevant information, while a longer one might still be meaningful if it includes necessary or supporting information \cite{kulesza2023too}. Second, the robustness analysis was limited to three types of augmentations: rotation by 10°, rotation by 180°, and Gaussian noise. While they reflect global and local image changes, other relevant transformations, like occlusion, color shifts, or brightness \cite{9176802,finzel2024near}, could be considered in the future. Thus, conclusions about the model’s robustness may not generalize to all forms of input variation. Additionally, the analysis focused on statistical changes in explanation length, but did not examine which concepts were added, removed, or altered under augmentation. This limits the interpretability of changes and assessment of whether the model shifted to semantically meaningful alternative concepts or just produced noisier explanations. However, this open question may be an interesting future evaluation step, especially if examined on richer data sets (e.g., showing the same object from different angles) and extended by a human study to measure understanding in explainees.

\section{Conclusion}\label{sec:conclusion}

This work introduced a concept-based contrastive explanation method for image classification, aiming to clarify why a model prefers one class over another using human-understandable concepts. Two key aspects were examined: the relationship between explanation length and concept relevance, and the robustness of explanations under image augmentations.

Experiments showed that highly relevant concepts led to shorter, more concise contrastive explanations, while less relevant ones produced longer, more diffuse outputs. This finding supports our hypothesis that relevance strength significantly influences explanation complexity, and that contrastive approaches can benefit from exploiting this relationship.

Robustness analysis revealed that explanation length remained stable under strong global transformations (e.g., 180° rotation), but changed notably with small local perturbations (e.g., 10° rotation, Gaussian noise), particularly within concept relevance ranges. This suggests that while explanations exhibit invariance to coarse transformations, the model is sensitive to fine-grained input variations, underscoring the role of geometrical properties of concepts.

Future extensions could leverage the finding that contrastive explanations, grounded in instance similarity and concept relevance, offer interpretable class distinctions. Several promising directions arise from this insight.

Models could be fine-tuned to classify instances with concept inputs disabled when those concepts were deemed irrelevant in a contrastive explanation task (i.e., mistakenly identified as important) \cite{finzel2024near,finzel2024relation}, potentially reducing \textit{Clever-Hans behavior} (see, e.g., Hernández-Orallo \cite{orallo2019gazing}) and improving generalization.

Rather than focusing solely on explanation length, future work could examine how individual concepts change under perturbations, identifying additions, removals, or alterations and their semantic relevance. Robustness testing could be expanded to include more realistic perturbations to better reflect real-world scenarios. Further, alternative explanation quality metrics like faithfulness, completeness, and user-perceived interpretability could be introduced and assessed via user studies \cite{finzel2024relation,schwalbe2023comprehensive} to better evaluate our method’s practical utility.

Overall, this work contributes to explainable AI by providing a concept-based contrastive explanation method for image classification, laying the groundwork for future research on explanation complexity and robustness.

%
%
%
\bibliographystyle{splncs04}
\bibliography{mybibliography}

\begin{thebibliography}{10}
\providecommand{\url}[1]{\texttt{#1}}
\providecommand{\urlprefix}{URL }
\providecommand{\doi}[1]{https://doi.org/#1}

\bibitem{achtibat2023attribution}
Achtibat, R., Dreyer, M., Eisenbraun, I., Bosse, S., Wiegand, T., Samek, W.,
  Lapuschkin, S.: From attribution maps to human-understandable explanations
  through concept relevance propagation. Nat. Mach. Intell.  \textbf{5}(9),
  1006--1019 (2023). \doi{https://doi.org/10.1038/s42256-023-00711-8}

\bibitem{adadi2018peeking}
Adadi, A., Berrada, M.: Peeking inside the black-box: {A} survey on explainable
  artificial intelligence {(XAI)}. {IEEE} Access  \textbf{6},  52138--52160
  (2018). \doi{10.1109/ACCESS.2018.2870052}

\bibitem{ali2023explainable}
Ali, S., Abuhmed, T., El{-}Sappagh, S.H.A., Muhammad, K., Alonso{-}Moral, J.M.,
  Confalonieri, R., Guidotti, R., Ser, J.D., Rodr{\'{\i}}guez, N.D., Herrera,
  F.: Explainable artificial intelligence {(XAI):} what we know and what is
  left to attain trustworthy artificial intelligence. Inf. Fusion  \textbf{99},
   101805 (2023). \doi{10.1016/J.INFFUS.2023.101805}

\bibitem{9176802}
Arcaini, P., Bombarda, A., Bonfanti, S., Gargantini, A.: Dealing with
  robustness of convolutional neural networks for image classification. In:
  2020 IEEE International Conference On Artificial Intelligence Testing
  (AITest). pp. 7--14 (2020). \doi{10.1109/AITEST49225.2020.00009}

\bibitem{bruckert2020roadmap}
Bruckert, S., Finzel, B., Schmid, U.: The next generation of medical decision
  support: {A} roadmap toward transparent expert companions. Frontiers Artif.
  Intell.  \textbf{3},  507973 (2020). \doi{10.3389/FRAI.2020.507973}

\bibitem{deng2009imagenet}
Deng, J., Dong, W., Socher, R., Li, L., Li, K., Fei{-}Fei, L.: Imagenet: {A}
  large-scale hierarchical image database. In: 2009 {IEEE} Computer Society
  Conference on Computer Vision and Pattern Recognition {(CVPR} 2009), 20-25
  June 2009, Miami, Florida, {USA}. pp. 248--255. {IEEE} Computer Society
  (2009). \doi{10.1109/CVPR.2009.5206848}

\bibitem{dhurandhar2018explanations}
Dhurandhar, A., Chen, P., Luss, R., Tu, C., Ting, P., Shanmugam, K., Das, P.:
  Explanations based on the missing: Towards contrastive explanations with
  pertinent negatives. In: Bengio, S., Wallach, H.M., Larochelle, H., Grauman,
  K., Cesa{-}Bianchi, N., Garnett, R. (eds.) Advances in Neural Information
  Processing Systems 31: Annual Conference on Neural Information Processing
  Systems 2018, NeurIPS 2018, December 3-8, 2018, Montr{\'{e}}al, Canada. pp.
  590--601 (2018)

\bibitem{doshivelez2017rigorous}
Doshi-Velez, F., Kim, B.: Towards a rigorous science of interpretable machine
  learning (2017), \url{https://arxiv.org/abs/1702.08608}

\bibitem{finzel2024relation}
Finzel, B., Hilme, P., Rabold, J., Schmid, U.: Telling more with concepts and
  relations: Exploring and evaluating classifier decisions with {CoReX}. arXiv
  preprint arXiv:2405.01661  (2024), \url{https://arxiv.org/abs/2405.01661}

\bibitem{finzel2024near}
Finzel, B., Knoblach, J., Thaler, A.M., Schmid, U.: Near hit and near miss
  example explanations for model revision in binary image classification. In:
  Juli{\'{a}}n, V., Camacho, D., Yin, H., Alberola, J.M., Nogueira, V.B.,
  Novais, P., Tall{\'{o}}n{-}Ballesteros, A.J. (eds.) Intelligent Data
  Engineering and Automated Learning - {IDEAL} 2024 - 25th International
  Conference, Valencia, Spain, November 20-22, 2024, Proceedings, Part {II}.
  Lecture Notes in Computer Science, vol. 15347, pp. 260--271. Springer (2024).
  \doi{10.1007/978-3-031-77738-7\_22}

\bibitem{fisher1992statistical}
Fisher, R.A.: Statistical Methods for Research Workers, pp. 66--70. Springer
  New York, New York, NY (1992). \doi{10.1007/978-1-4612-4380-9\_6}

\bibitem{gunning2019darpa}
Gunning, D., Aha, D.W.: Darpa's explainable artificial intelligence {(XAI)}
  program. {AI} Mag.  \textbf{40}(2),  44--58 (2019).
  \doi{10.1609/AIMAG.V40I2.2850}

\bibitem{orallo2019gazing}
Hern{\'{a}}ndez{-}Orallo, J.: Gazing into clever hans machines. Nat. Mach.
  Intell.  \textbf{1}(4),  172--173 (2019). \doi{10.1038/S42256-019-0032-5}

\bibitem{kulesza2023too}
Kulesza, T., Stumpf, S., Burnett, M.M., Yang, S., Kwan, I., Wong, W.: Too much,
  too little, or just right? ways explanations impact end users' mental models.
  In: Kelleher, C., Burnett, M.M., Sauer, S. (eds.) 2013 {IEEE} Symposium on
  Visual Languages and Human Centric Computing, San Jose, CA, USA, September
  15-19, 2013. pp. 3--10. {IEEE} Computer Society (2013).
  \doi{10.1109/VLHCC.2013.6645235}

\bibitem{vanLent2004explainable}
van Lent, M., Fisher, W., Mancuso, M.: An explainable artificial intelligence
  system for small-unit tactical behavior. In: Proc. 2004 Nat. Conf. Artificial
  Intelligence. pp. 900--907. AAAI Press; MIT Press, 2004

\bibitem{lipton1990contrastive}
Lipton, P.: Contrastive explanation. Royal Institute of Philosophy Supplement
  \textbf{27},  247–266 (1990). \doi{10.1017/S1358246100005130}

\bibitem{margolis2007ontology}
Margolis, E., Laurence, S.: The ontology of concepts—abstract objects or
  mental representations? Noûs  \textbf{41}(4),  561--593 (2007).
  \doi{https://doi.org/10.1111/j.1468-0068.2007.00663.x}

\bibitem{miller1995wordnet}
Miller, G.A.: {WordNet}: {A} lexical database for english. Commun. {ACM}
  \textbf{38}(11),  39--41 (1995). \doi{10.1145/219717.219748}

\bibitem{miller2019explanation}
Miller, T.: Explanation in artificial intelligence: Insights from the social
  sciences. Artif. Intell.  \textbf{267},  1--38 (2019).
  \doi{10.1016/J.ARTINT.2018.07.007}

\bibitem{palmer2002psychological}
Palmer, D.C.: Psychological essentialism: A review of {E. Margolis and S.
  Laurence} (eds.), {C}oncepts: Core readings (2002)

\bibitem{poeta2023concept}
Poeta, E., Ciravegna, G., Pastor, E., Cerquitelli, T., Baralis, E.:
  Concept-based explainable artificial intelligence: A survey  (2023),
  \url{https://arxiv.org/abs/2312.12936}

\bibitem{prabhushankar2020contrastive}
Prabhushankar, M., Kwon, G., Temel, D., AlRegib, G.: Contrastive explanations
  in neural networks. In: 2020 IEEE International Conference on Image
  Processing (ICIP). pp. 3289--3293. IEEE (2020)

\bibitem{rabold2022nearmisses}
Rabold, J., Siebers, M., Schmid, U.: Generating contrastive explanations for
  inductive logic programming based on a near miss approach. Machine Learning
  \textbf{111}(5),  1799--1820 (2022). \doi{10.1007/s10994-021-06048-w}

\bibitem{rudin2019stop}
Rudin, C.: Stop explaining black box machine learning models for high stakes
  decisions and use interpretable models instead. Nat. Mach. Intell.
  \textbf{1}(5),  206--215 (2019). \doi{10.1038/S42256-019-0048-X}

\bibitem{ILSVRC15}
Russakovsky, O., Deng, J., Su, H., Krause, J., Satheesh, S., Ma, S., Huang, Z.,
  Karpathy, A., Khosla, A., Bernstein, M., Berg, A.C., Fei-Fei, L.: {ImageNet
  Large Scale Visual Recognition Challenge}. International Journal of Computer
  Vision (IJCV)  \textbf{115}(3),  211--252 (2015).
  \doi{10.1007/s11263-015-0816-y}

\bibitem{schwalbe2023comprehensive}
Schwalbe, G., Finzel, B.: A comprehensive taxonomy for explainable artificial
  intelligence: a systematic survey of surveys on methods and concepts. Data
  Min. Knowl. Discov.  \textbf{38}(5),  3043--3101 (2024).
  \doi{10.1007/S10618-022-00867-8}

\bibitem{simonyan2014very}
Simonyan, K., Zisserman, A.: Very deep convolutional networks for large-scale
  image recognition. In: Bengio, Y., LeCun, Y. (eds.) 3rd International
  Conference on Learning Representations, {ICLR} 2015, San Diego, CA, USA, May
  7-9, 2015, Conference Track Proceedings (2015),
  \url{http://arxiv.org/abs/1409.1556}

\bibitem{stepin2021survey}
Stepin, I., Alonso, J.M., Catal{\'{a}}, A., Pereira{-}Fari{\~{n}}a, M.: A
  survey of contrastive and counterfactual explanation generation methods for
  explainable artificial intelligence. {IEEE} Access  \textbf{9},  11974--12001
  (2021). \doi{10.1109/ACCESS.2021.3051315}

\bibitem{student1908probable}
Student: The probable error of a mean. Biometrika  \textbf{6}(1),  1--25
  (1908). \doi{10.2307/2331554}

\bibitem{winston1975learning}
Winston, P.H.: Learning structural descriptions from examples. In: The
  Psychology of Computer Vision, pp. 157–--210. McGraw-Hill (1975)

\end{thebibliography}
\newpage

\appendix
\section{Appendix}

\begin{figure}
\centering
\subfloat[Very strong relevance\label{fig:rot-verystong}]{
    \centering 
    \includegraphics[width=0.45\textwidth, alt={Box plot comparing explanation lengths for "Original" and "Rotated" images under the category "Very Strong Relevance." The y-axis represents explanation length, ranging from 5 to 30. Both plots show median, quartiles, and outliers, with the "Original" having a slightly wider interquartile range.}]{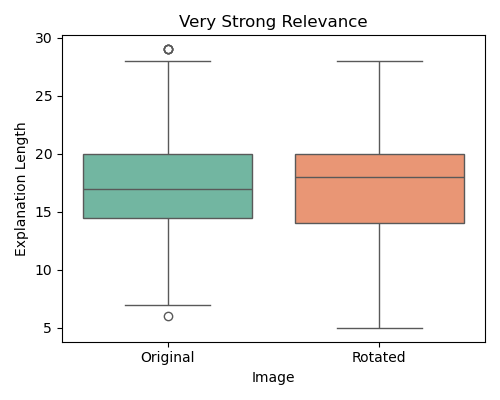}}

\hfill 

\subfloat[Strong relevance\label{fig:rot-strong}]{
    \centering 
    \includegraphics[width=0.45\textwidth, alt={Box plot comparing explanation lengths for "Original" and "Rotated" images under the title "Strong Relevance." The y-axis represents explanation length, ranging from 20 to 60. Both plots show similar median values around 40, with some outliers.}]{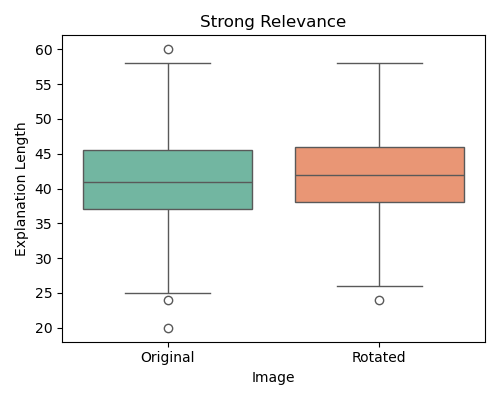}}

\subfloat[Low relevance\label{fig:rot-low}]{
    \centering 
    \includegraphics[width=0.45\textwidth, alt={Box plot comparing explanation lengths for "Original" and "Rotated" images under the category "Low Relevance." The y-axis represents explanation length, ranging from 50 to 100. Both plots show median, quartiles, and outliers, with similar distributions.}]{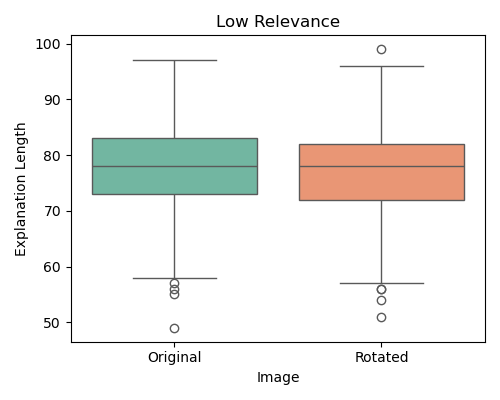}}

\hfill 

\subfloat[Very low relevance\label{fig:rot-verylow}]{
    \centering 
    \includegraphics[width=0.45\textwidth, alt={Box plot comparing explanation lengths for "Original" and "Rotated" images under the category "Very Low Relevance." The y-axis represents explanation length, ranging from 70 to 130. Both plots show similar median values around 100, with "Original" having a slightly wider interquartile range and a few outliers below 80.}]{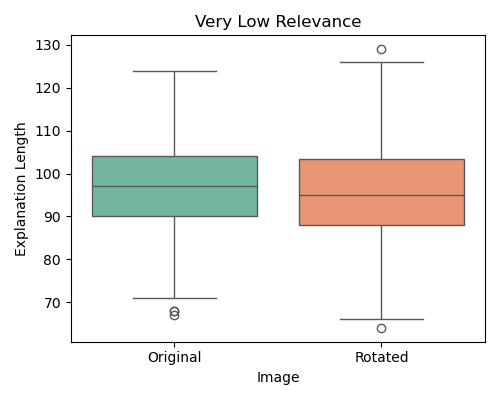}}

\caption{Explanation length for Original vs. 180-degree Rotation (n = 615 teapots, 328 vases)}
\label{fig:rot}
\end{figure}

\begin{figure}
\centering
\subfloat[Very strong relevance\label{fig:rot10-verystong}]{
    \centering 
    \includegraphics[width=0.45\textwidth, alt={Box plot comparing explanation lengths for "Original" and "Rotation" images under the category "Very Strong Relevance." The y-axis represents explanation length, ranging from 5 to 30. Both plots show median lines, interquartile ranges, and outliers.}]{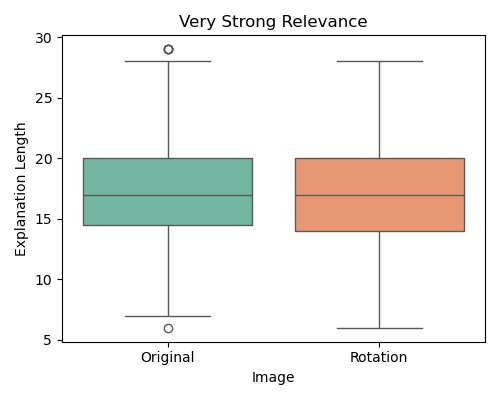}}

\hfill

\subfloat[Strong relevance\label{fig:rot10-strong}]{
    \centering 
    \includegraphics[width=0.45\textwidth, alt={Box plot comparing explanation lengths for "Original" and "Rotation" images under the title "Strong Relevance." The y-axis represents explanation length, ranging from 20 to 60. Both plots show median values around 40, with similar interquartile ranges. Outliers are present in the "Original" category.}]{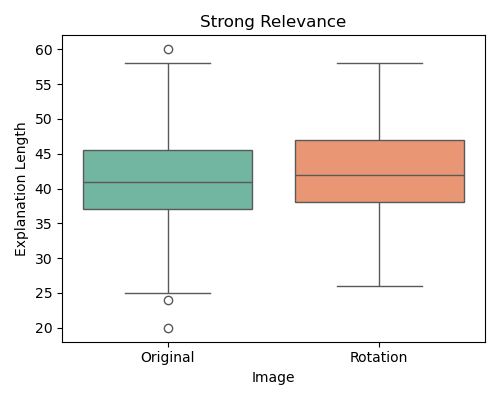}}

\subfloat[Low relevance\label{fig:rot10-low}]{
    \centering 
    \includegraphics[width=0.45\textwidth, alt={Box plot titled "Low Relevance" comparing explanation lengths for "Original" and "Rotation" images. Both plots show median values around 80, with interquartile ranges from approximately 70 to 90. The "Original" plot has several outliers below 60, while the "Rotation" plot has one outlier below 60. The y-axis is labeled "Explanation Length."}]{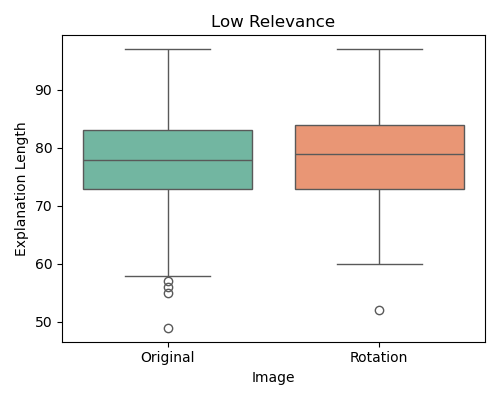}}

\hfill 

\subfloat[Very low relevance\label{fig:rot10-verylow}]{
    \centering 
    \includegraphics[width=0.45\textwidth, alt={Box plot comparing explanation lengths for "Original" and "Rotation" images under the category "Very Low Relevance." The y-axis represents explanation length, ranging from 70 to 130. Both box plots show similar median values around 100, with slight variations in interquartile ranges and outliers.}]{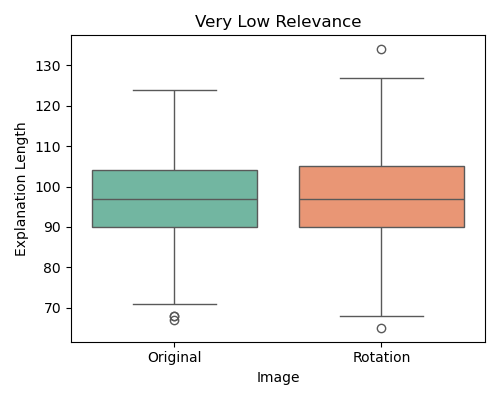}}

\caption{Explanation length for Original vs. 10-degree Rotation (n = 615 teapots, 328 vases)}
\label{fig:rot10}
\end{figure}

\begin{figure}
\centering
\subfloat[Very strong relevance\label{fig:gausnois-verystong}]{
    \centering 
    \includegraphics[width=0.45\textwidth, alt={Box plot comparing explanation lengths for "Original" and "Gaussian Noise" images under the title "Very Strong Relevance." The plot shows median, quartiles, and outliers for each category, with "Original" having a higher median explanation length than "Gaussian Noise." The y-axis is labeled "Explanation Length," ranging from 0 to 30.}]{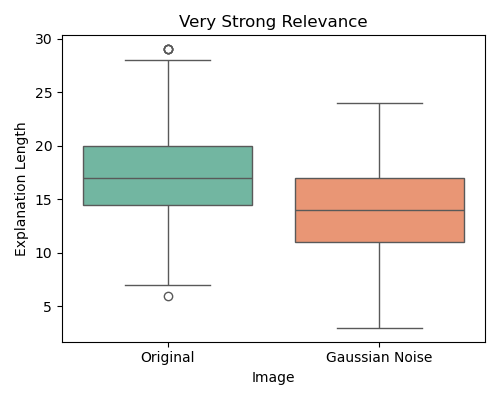}}

\hfill 

\subfloat[Strong relevance\label{fig:gausnois-strong}]{
    \centering 
    \includegraphics[width=0.45\textwidth, alt={Box plot comparing explanation lengths for "Original" and "Gaussian Noise" under the title "Strong Relevance." The y-axis represents explanation length, ranging from 20 to 60. The "Original" box plot shows a median around 40, with outliers above 60 and below 20. The "Gaussian Noise" box plot has a median slightly below 40, with outliers below 20.}]{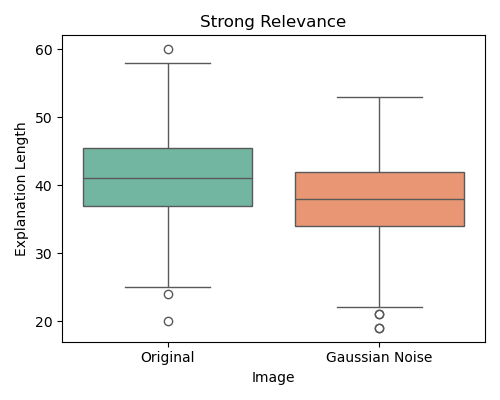}}

\subfloat[Low relevance\label{fig:gausnois-low}]{
    \centering 
    \includegraphics[width=0.45\textwidth, alt={Box plot titled "Low Relevance" comparing explanation lengths for "Original" and "Gaussian Noise" images. The y-axis represents explanation length, ranging from 50 to 90. Both plots show median, quartiles, and outliers, with "Original" having a slightly higher median than "Gaussian Noise."}]{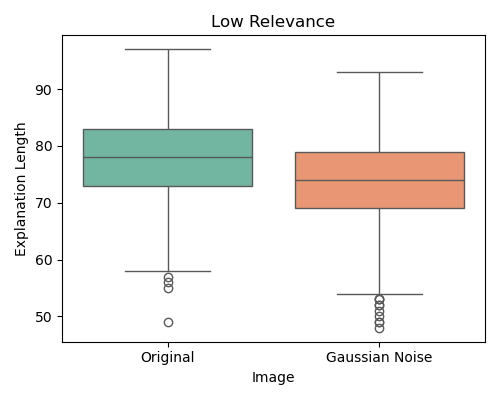}}

\hfill 

\subfloat[Very low relevance\label{fig:gausnois-verylow}]{
    \centering 
    \includegraphics[width=0.45\textwidth, alt={Box plot comparing explanation lengths for "Original" and "Gaussian Noise" images under the title "Very Low Relevance." The y-axis represents explanation length, ranging from 50 to 120. The "Original" image has a median around 100, with a range from approximately 70 to 120, and some outliers below 70. The "Gaussian Noise" image has a median around 90, with a range from approximately 60 to 110, and outliers below 60.}]{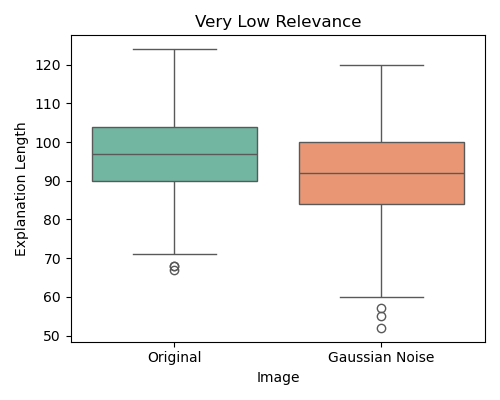}}

\caption{Explanation length for Original vs. Gaussian Noise (n = 615 teapots, 328 vases)}
\label{fig:gausnois}
\end{figure}

\end{document}